\documentclass[twocolumn,10pt]{article}

\usepackage{graphicx}
\usepackage{subfigure}
\usepackage{natbib}
\usepackage{amsmath}

\graphicspath{{imgs/}}

\twocolumn  
\newlength\titlebox 

\title{Matching Objects across the Textured--Smooth Continuum}
\author{
Ognjen Arandjelovi\'c\\
Centre for Pattern Recognition and Data Analytics\\
Deakin University\\
Geelong VIC 3220\\
Australia\\
ognjen.arandjelovic@gmail.com}

\date{}

\begin{document}

\maketitle

\thispagestyle{empty}

\begin{abstract}
The problem of 3D object recognition is of immense practical importance, with the last decade witnessing a number of breakthroughs in the state of the art. Most of the previous work has focused on the matching of textured objects using local appearance descriptors extracted around salient image points. The recently proposed bag of boundaries method was the first to address directly the problem of matching smooth objects using boundary features. However, no previous work has attempted to achieve a holistic treatment of the problem by jointly using textural and shape features which is what we describe herein. Due to the complementarity of the two modalities, we fuse the corresponding matching scores and learn their relative weighting in a data specific manner by optimizing discriminative performance on synthetically distorted data. For the textural description of an object we adopt a representation in the form of a histogram of SIFT based visual words. Similarly the apparent shape of an object is represented by a histogram of discretized features capturing local shape. On a large public database of a diverse set of objects, the proposed method is shown to outperform significantly both purely textural and purely shape based approaches for matching across viewpoint variation.
\end{abstract}

\vspace{-6pt}\section{Introduction}\label{s:intro}
\begin{figure}[htb]
  \centering
  \subfigure[\tiny Textured]{\includegraphics[width=0.14\textwidth]{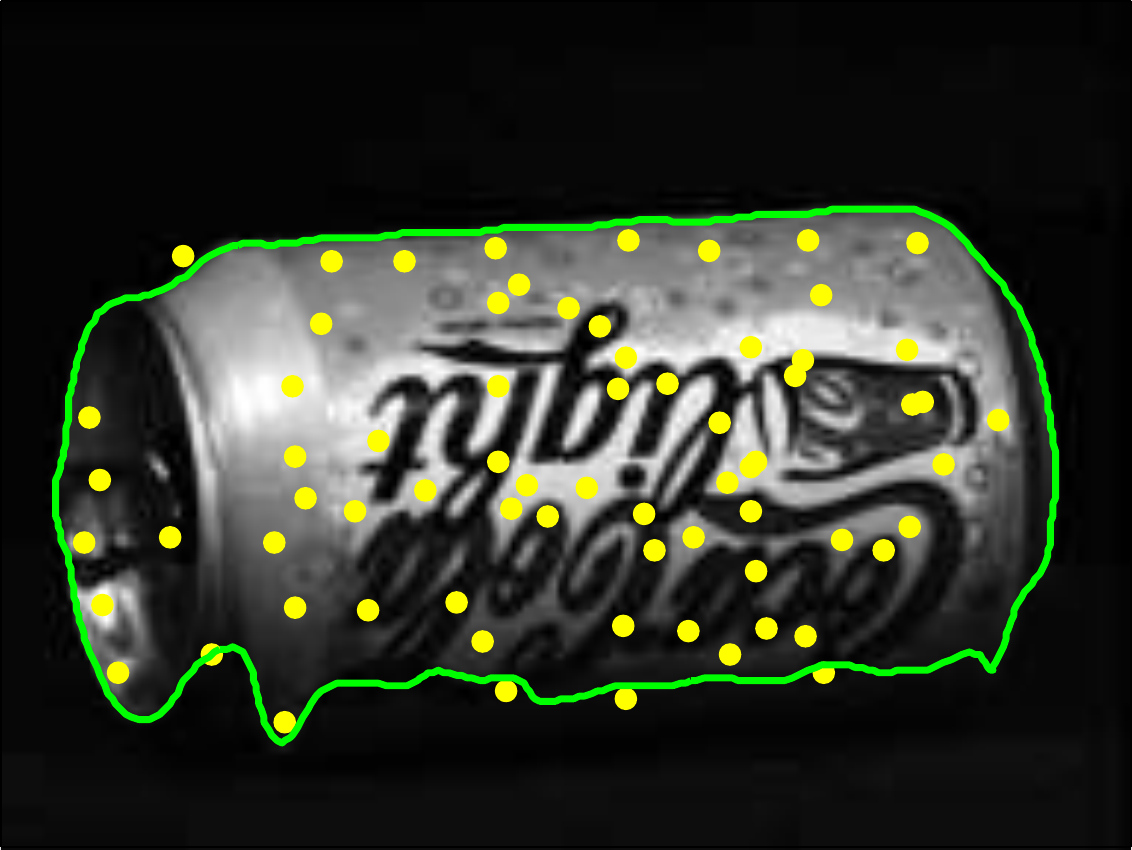}\label{f:objT}}\hspace{10pt}
  \subfigure[\tiny Untextured]{\includegraphics[width=0.14\textwidth]{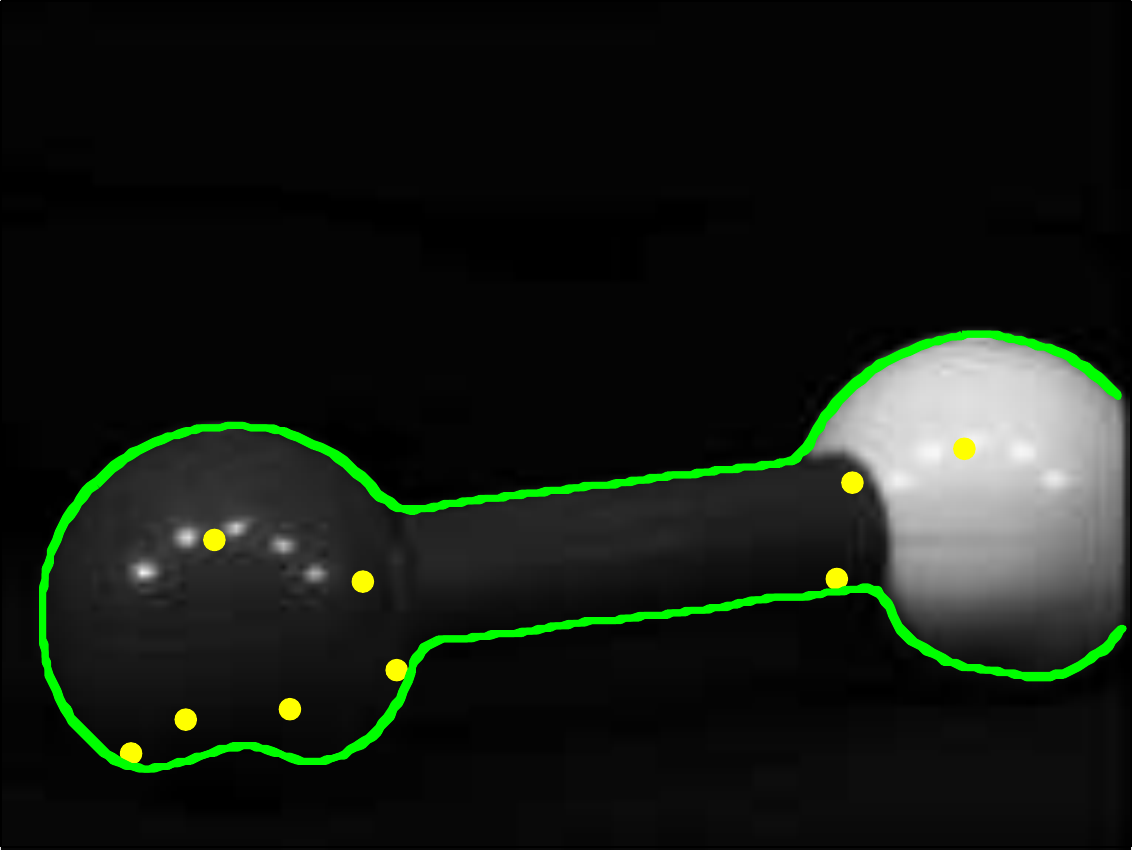}\label{f:objS}}\hspace{10pt}
  \subfigure[\tiny Semi-textured]{\includegraphics[width=0.14\textwidth]{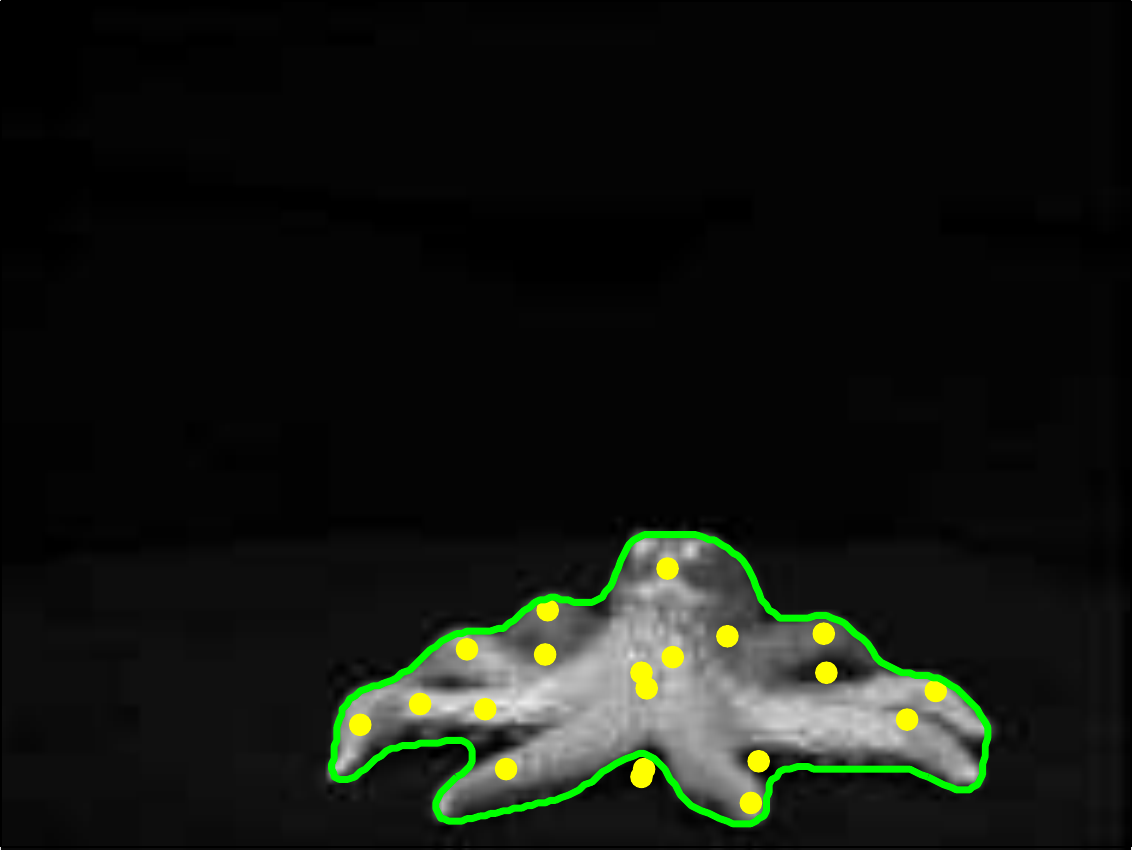}\label{f:objG}}
  \caption{ Most previous work on object recognition and retrieval focused on textured objects, such as that shown in (a). These objects are characterized by large numbers of interest points (shown as yellow dots) which correspond to regions of discriminative local appearance. In contrast, the recent work by Arandjelovi\'c and Zisserman specifically deals with smooth objects, such as in that shown in (b), which produce few interest points. The characteristic appearance of these objects must be described in terms of their apparent shape (shown as a green line). However, many objects, such as that in (c), can neither be considered textured nor smooth and neither of the previously proposed representations capture their appearance sufficiently well. }
\end{figure}

The problem of recognizing 3D objects from images has been one of the most active areas of computer vision research in the last decade. This is a consequence not only of the high practical potential of automatic object recognition systems but also of significant breakthroughs which have facilitated the development of fast and reliable solutions \citep{SiviZiss2003,Lowe2004,SiviRussEfro+2005,NistStew2006}. These mainly centre around the detection of robust and salient image loci (keypoints) or regions~\citep{Lowe2004,MikoSchm2004} and the characterization of their appearance using local descriptors~\citep{Lowe2004,MikoTuytSchmZiss+2005}. While highly successful in the recognition of textured objects even in the presence of significant viewpoint and scale changes, these methods fail when applied on texturally smooth (i.e.\ nearly textureless) objects~\citep{AranZiss2011}. Unlike textured objects such as that in Figure~\ref{f:objT}, smooth objects inherently do not exhibit appearance from which well localized keypoints and thus discriminative local descriptors can be extracted. This can be readily observed on the example in Figure~\ref{f:objS}. The failure of keypoint based methods in adequately describing the appearance of smooth objects has recently been convincingly demonstrated using images of sculptures~\citep{AranZiss2011}.

\begin{figure*}[htb]
  \centering
  \includegraphics[width=1\textwidth]{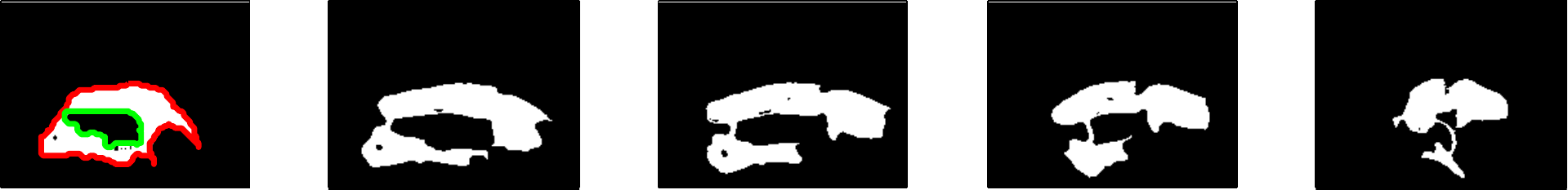}
  \caption{ As seen on the example in this figure (the second object from the Amsterdam Library of Object Images \citep{GeusBurgSmeu2005}), the apparent shape of 3D objects changes dramatically with viewpoint. Matching is made even more difficult by errors introduced during automatic segmentation. The leftmost image in the figure also shows automatically delineated object boundaries -- one external boundary is shown in red and one internal boundary in green. }
  \label{f:segmentation}
\end{figure*}

\paragraph{Smooth objects.} Considering that their texture is not informative, characteristic discriminative information of smooth objects must be extracted from shape instead. However, it is extremely difficult to formulate a meaningful prior which would allow the reconstruction of an accurate depth map from a single image of a general 3D object; consequently, matching must be performed based on the objects' \emph{apparent} shape as it is observed in images. This is a most difficult task because apparent shape is greatly affected by out of plane rotation of the object. What is more, the extracted shape is likely to contain errors when the object is automatically segmented out from realistic, cluttered images. These challenges, illustrated in Figure~\ref{f:segmentation}, were first addressed by the bag of boundaries (BoB) method of \cite{AranZiss2011}.

At this point it is worth emphasizing the difference between the problem we tackle in this paper and that addressed by \cite{FerrTuytVanG2006,SavaFeiF2007,LuLateAdluLing+2009,SrinZhuShi2010}, and \cite{PayeTodo2011}, amongst others. This corpus of previous work aims at detecting an object of a \emph{specific class} while our goal here is to recognize a \emph{particular object instance}. Object class detection methods usually discriminate between a small number of classes (5--10), use a large number of training images per class (hundreds of images is not uncommon) to learn the scope of possible appearance variation, and usually assume that the pose of different objects in a class is similar (or, equivalently, explicitly include different poses of interest in training, with little direct attention to generalization to unseen poses). In contrast, we are interested in discriminating between a large number of specific objects (1000) across pose/viewpoint, using only a single image per object both for training and as unknown queries.

\begin{figure}[htb]
  \centering
  \subfigure[Textural features]{\includegraphics[width=0.35\textwidth]{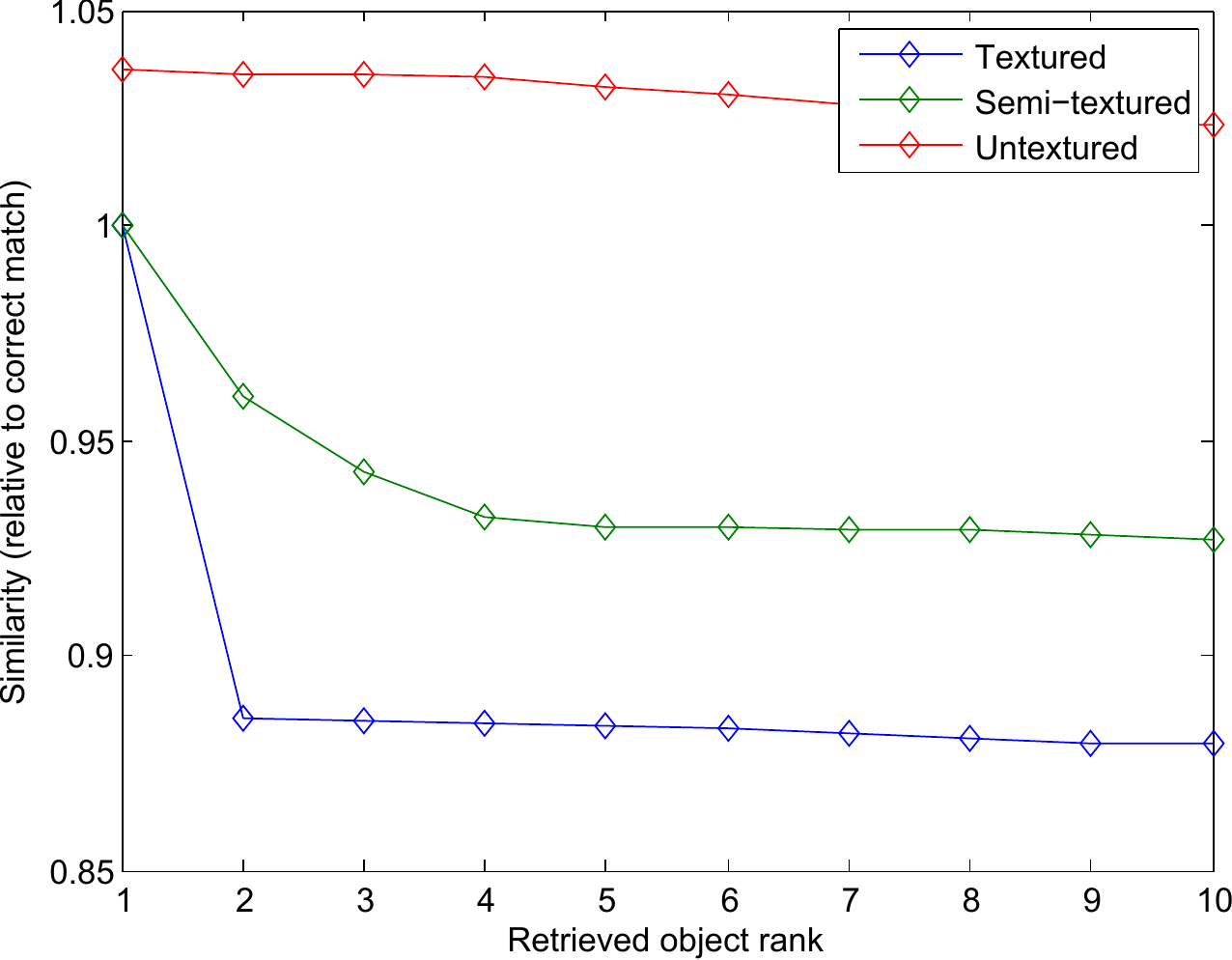}\label{f:objcomp1}}
  \subfigure[Contour features]{\includegraphics[width=0.35\textwidth]{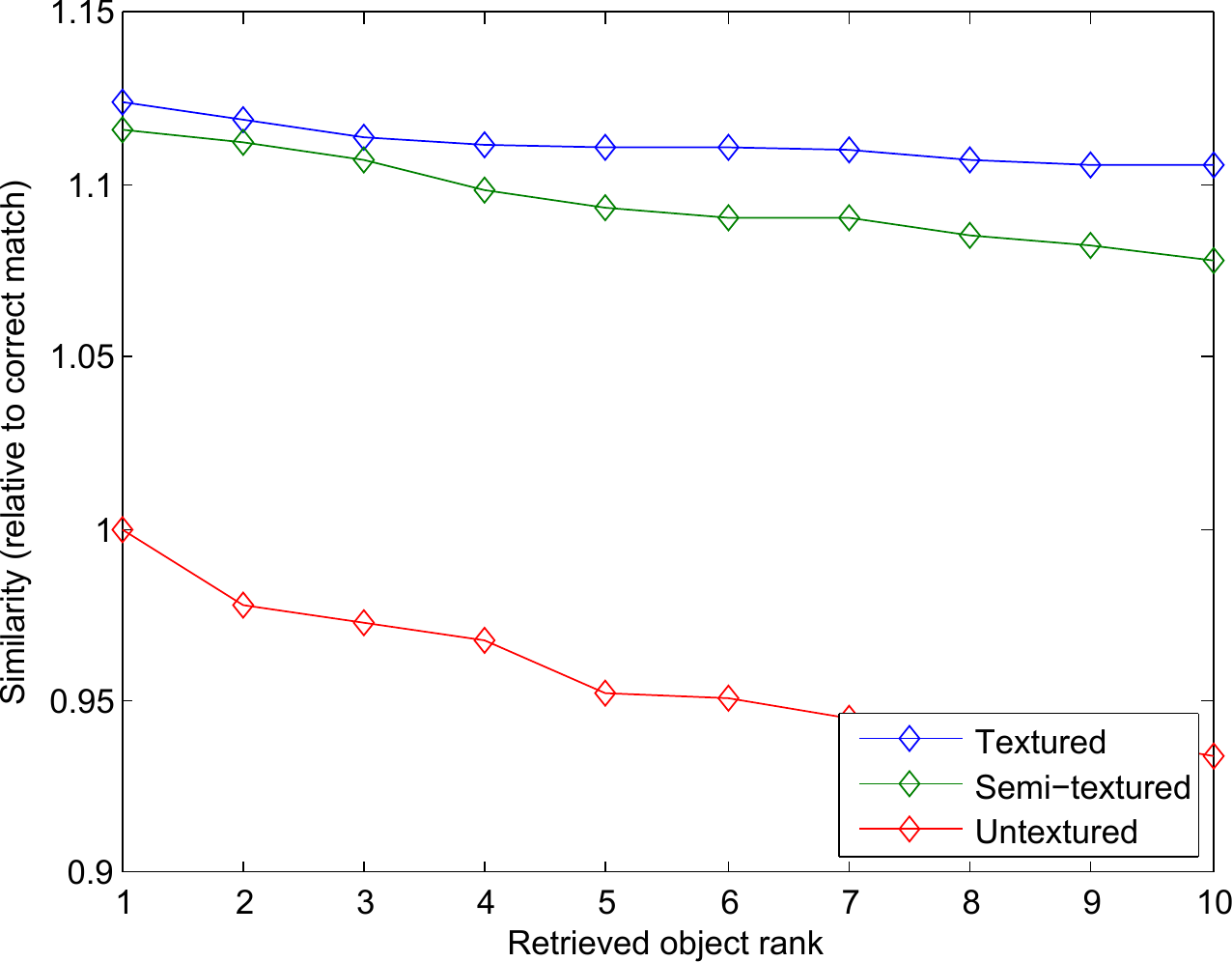}\label{f:objcomp2}}
  \caption{ Similarity scores of the top 10 object matches in the database for the three query objects in Figure~1 (textured, semi-textured and smooth), using features based on (a) texture and (b) shape only. The database contains objects seen from a nominal 0$^\circ$ viewpoint while querying was performed after a 30$^\circ$ yaw change. Similarity scores are scaled relative to the correct match (e.g.\ the matching score of 1.0 for the top ranking object means that the correct image was retrieved first). }
\end{figure}

\paragraph{General objects.} All of the previous work on the matching of 3D objects considers objects to be either textured or smooth. However, both of these attributes are exhibited across a continuum: different objects are characterized by varying extents of texturedness and smoothness. Thus the key premise of the present work is that both texture and shape based descriptors should be used to represent the appearance of an object. In addition, the relative significance of texture and shape should be learnt in a data specific manner -- for some data sets texture is more discriminative than shape and \textit{vice versa} for others. This is illustrated in Figures~\ref{f:objcomp1} and~~\ref{f:objcomp2} which show the results of retrieval using textural and shape features (see Section~\ref{s:proposed}) for the three query objects in Figure~1(a--c). As expected, textural features perform better than shape for textured objects, while for smooth objects it is the other way round. For semi-textured objects, the performance achieved with either of the modalities lies somewhere between the two extremes.

\vspace{-6pt}\section{Proposed method}\label{s:proposed}{
The texture of an object and its apparent shape are virtually entirely complementary discriminative characteristics. Therefore we extract and process the representations of the two modalities independently. An object's texture is captured using a histogram computed over a vocabulary of textural words, learnt by clustering local texture descriptors extracted from the training data set. Similarly, a histogram over a vocabulary of elementary shapes, learnt by clustering local shape descriptors, is used to capture the object's shape. We adopt the standard SIFT descriptor as the basic building block of the texture representation and a recently introduced descriptor of local shape for the characterization of shape \cite{Aran2012f}. Unlike the implicit image-based descriptor at the centre of the BoB method, the shape descriptor used here is extracted directly from shape itself, following foreground(object)/background segmentation.

\vspace{-6pt}\subsection{Object shape representation}\label{ss:shape}
The first part of our characterization of an object is purely based on its apparent shape in an image. To emphasize the difference of the approach adopted here in comparison with previous work, we start with an overview of the only method in the literature which addressed the same problem. This is followed by a detailed description of the representation.

\vspace{-6pt}\subsubsection*{Previous work: bag of boundaries (BoB)}\vspace{-6pt}\label{sss:bob}
The bag of boundaries method of \cite{AranZiss2011} describes the apparent shape of an object using its boundaries, external as well as internal, as shown in Figure~\ref{f:segmentation}. The boundaries are traced and elementary descriptors extracted at equidistant loci. At each point at which descriptors are extracted, three descriptors of the same type are extracted at different scales, computed relative to the foreground object area.

\paragraph{Baseline descriptor.} The semi-local elementary descriptor is computed from the image patch centred at a boundary point and it consists of two parts. The first of these is similar to the histogram of oriented gradients (HoG) representation of appearance~\citep{DalaTrig2005}. Arandjelovi\'c and Zisserman compute a weighted histogram of gradient orientations for each $8\times 8$ pixel cell of the image patch which is resized to the uniform scale of $32 \times 32$ pixels, and concatenate these for each $3 \times 3$ cell region. These region descriptors are then concatenated themselves, resulting in a vector of dimension 324 (there are 4 regions, each with 9 cells and each cell is represented using a 9 direction histogram) which is $L_2$ normalized. The second part of the descriptor is what the authors term the occupancy matrix. The value of each element of this $4 \times 4$ matrix is the proportion of foreground (object) pixels in the corresponding region of the local patch extracted around the boundary. This matrix is rasterized, $L_2$ normalized and concatenated with the corresponding HoG vector to produce the final 340 dimension descriptor.

\paragraph{Matching.} Arandjelovi\'c and Zisserman apply their descriptor in the standard framework used for large scale retrieval. Firstly, the descriptor space is discretized by clustering the descriptors extracted from the entire data set. The original work used 10,000 clusters. Each object is then described by the histogram of the corresponding descriptor cluster memberships. This histogram is what the authors call a bag of boundaries. Note that the geometric relationship between different boundary descriptors is not encoded and that the descriptors extracted at different scales at the same boundary locus are bagged independently. Finally, retrieval ordering is determined by matching object histograms using the Euclidean distance following the usual \emph{tf-idf} weighting~\citep{WuLukWongKwok2008}.

\paragraph{Limitations.} The first major difference between the BoB method and that used in the present work lies in the manner in which boundary loci are selected. In BoB all segments of the boundary are treated with equal emphasis, extracting descriptors at equidistant points. However, not all parts of the boundary are equally informative. In addition, a dense representation of this type is inherently sensitive to segmentation errors even when they are in the non-discriminative regions of the boundary. Thus we adopt the use of a sparse representation which seeks to describe the shape of the boundary in the proximity of salient boundary loci only which are detected automatically. The second major difference between the BoB method and ours, is to be found in the form of the local boundary descriptor. The BoB descriptor is image based. The consideration of a wide image region, when it is only a characterization of the local boundary that is needed is not only inefficient, but as an explicit description also likely not the most robust or discriminative representation. In contrast, our boundary descriptor is explicitly based on local shape (see \cite{Aran2012f} for a more detailed discussion and comparative evaluation).

\vspace{-6pt}\subsubsection*{Boundary keypoint detection}\vspace{-6pt}\label{sss:keypt}
The problem of detecting characteristic image loci is well researched and a number of effective methods have been described in the literature; examples include approaches based on the difference of Gaussians~\citep{Lowe2004} and wavelet trees~\citep{FauqKingAnde2006}. When dealing with keypoints in images, the meaning of saliency naturally emerges as a property of appearance (pixel intensity) which is directly measured. This is not the case when dealing with curves for which saliency has to be defined by means of higher order variability which is computed rather than directly measured. As proposed by \cite{Aran2012f}}, in this paper we detect characteristic boundary loci as points of local curvature maxima, computed at different scales. Starting from the finest scale after localizing the corresponding keypoints, Gaussian smoothing is applied to the boundary which is then downsampled for the processing at a coarser scale. As recommended in the original publication, we adopt the granularity step size of 2, i.e.\ we downsample the boundary by one octave for processing at the next scale.

We estimate the curvature at the $i$-th vertex by the curvature of the circular arc fitted to three consecutive boundary vertices: $i-1$, $i$ and $i+1$. The method used to perform Gaussian smoothing of the boundary is summarized next.

\begin{figure*}[htb]
  \centering
  \subfigure[Contour]{\includegraphics[height=0.2\textwidth]{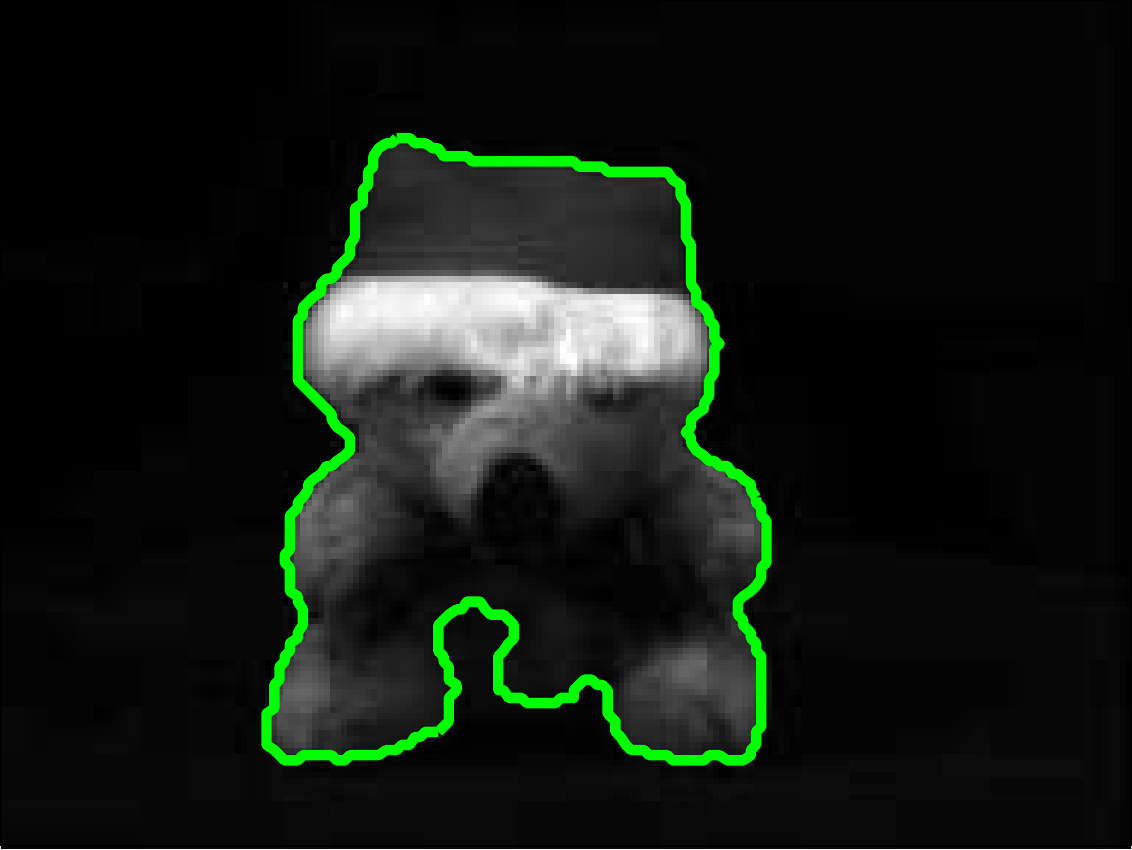}\label{f:contour}}~\hspace{30pt}
  \subfigure[Keypoints]{\includegraphics[height=0.2\textwidth]{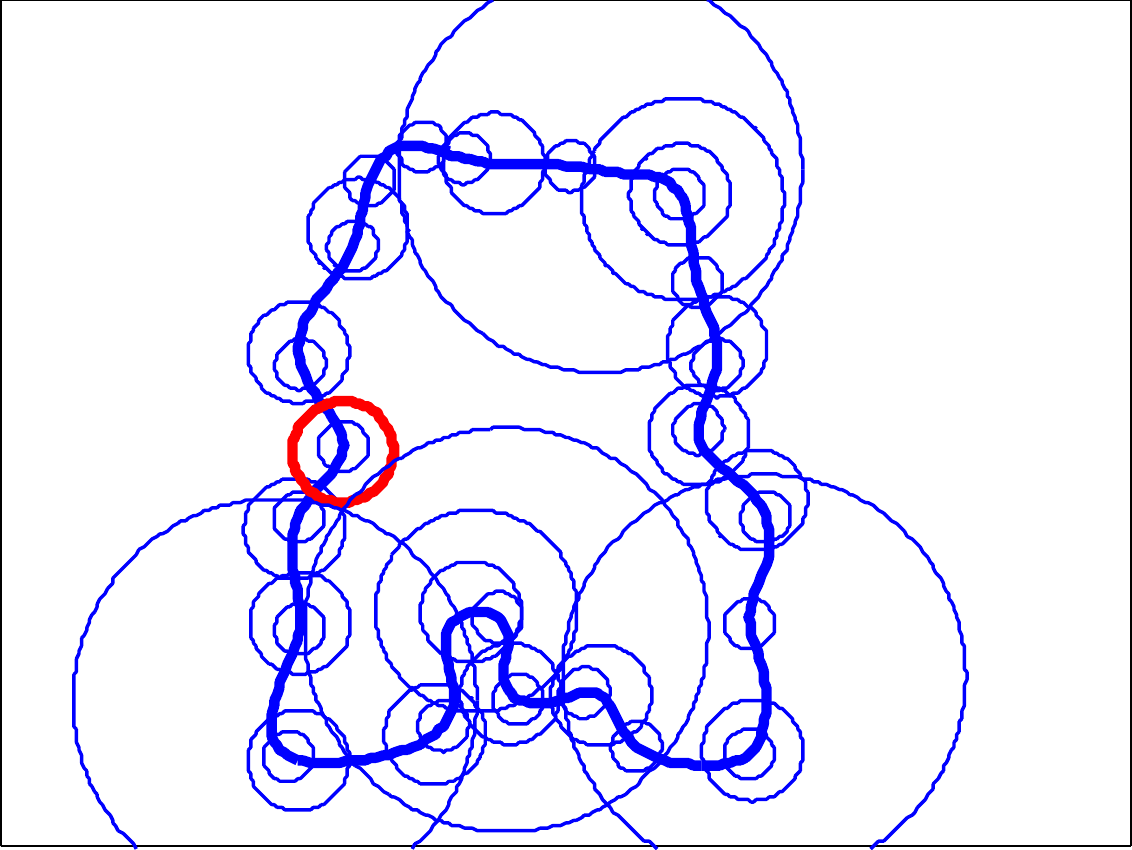}\label{f:keypts}}~\hspace{30pt}
  \subfigure[Descriptor]{\includegraphics[height=0.2\textwidth]{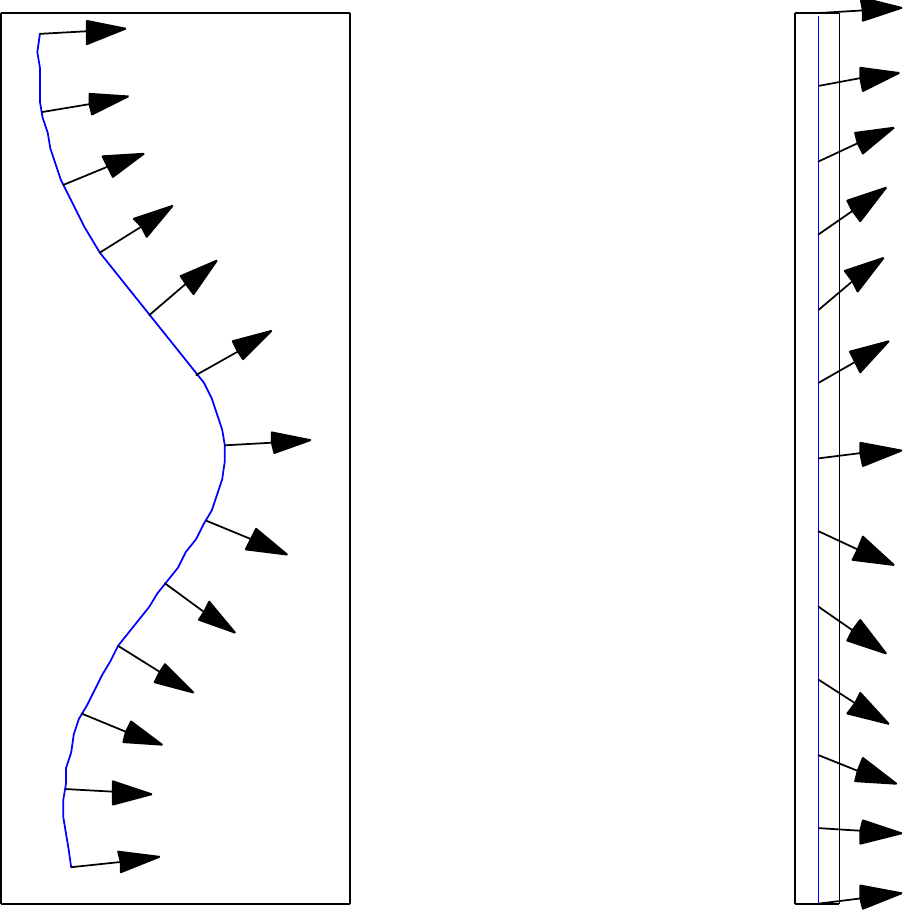}\label{f:desc}}
  \vspace{5pt}
  \caption{ (a) Original image of an object overlaid with the object boundary (green line), (b) the corresponding boundary keypoints detected using the method proposed in Section~\ref{sss:keypt} and (c) an illustration of a local boundary descriptor based on the profile of boundary normals' directions (the corresponding interest point is shown in red in (b)).  }
\end{figure*}

\paragraph{Boundary curve smoothing.}
A straightforward approach to smoothing a curve such as the object boundary is to replace each of its vertices $c_i$ (a 2D vector) by a Gaussian-weighted sum of vectors corresponding to its neighbours:
\begin{align}
  c'_i = \sum_{j=-w}^w G_j \times c_{i+j}
\end{align}
where $G_j$ is the $j$-th element of a Gaussian kernel with the width $2w+1$. This is analogous to the Gaussian smoothing of images. However, this method introduces an undesirable artefact which is demonstrated as a gradual shrinkage of the boundary. In the limit, repeated smoothing results in the collapse to a point -- the centre of gravity of the initial curve. We solve this problem using an approach inspired by Taubin's work~\citep{Taub1995}. The key idea is that \emph{two} smoothing operations are applied, with the second update to the boundary vertices in the ``negative'' direction. The second smoothing is applied on the results of the first:
\begin{align}
  c''_i = \sum_{j=-w}^w G_j \times c'_{i+j}.
\end{align}
The final result $\tilde{c}_i$ is computed by subtracting the differential of the second smoothing from the result of the first smoothing, weighted by a positive constant $K$:
\begin{align}
  \tilde{c}_i = c'_i - K \times (c''_i - c'_i).
\end{align}
The constant $K$ is determined by requiring that in the limit, repeated smoothing does not change the circumference of the boundary. In other words, repeated smoothing should cause the boundary to converge towards a circle of the radius $l_c/(2\pi)$ where $l_c$ is the circumference of the initial boundary. For this to be the case, smoothing should leave the aforesaid circle unaffected. It can be shown that this is satisfied iff:
\begin{align}
  K = \Big[ \sum_{j=-w}^w G_j \times \cos(j \phi) \Big]^{-1}
\end{align}
where $\phi = 2\pi/n_v$ and $n_v$ is the number of boundary vertices. An example of a boundary contour and the corresponding interest point loci are shown respectively in Figures~\ref{f:contour} and~\ref{f:keypts}.

\vspace{-6pt}\subsubsection*{Local boundary descriptor}\vspace{-6pt}\label{sss:shpdesc}
Following the detection of boundary keypoints, our goal is to describe the local shape of the boundary. After experimenting with a variety of descriptors based on local curvatures, angles and normals, using histogram and order preserving representations, we found that the best results are achieved using a local profile of boundary normals' directions.

To extract a descriptor, we sample the boundary around a keypoint's neighbourhood (at the characteristic scale of the keypoint) at $n_s$ equidistant points and estimate the boundary normals' directions at the sampling loci. This is illustrated in Figure~\ref{f:desc}. Boundary normals are estimated in a similar manner as curvature in Section~\ref{sss:keypt}. For each sampling point, a circular arc is fitted to the closest boundary vertex and its two neighbours, after which the desired normal is approximated by the corresponding normal of the arc, computed analytically. The normals are scaled to unit length and concatenated into the final descriptor with $2 n_s$ dimensions. After experimenting with different numbers of samples, from as few as 4 up to 36, we found that our method exhibited little sensitivity to the exact value of this parameter. For the experiments in this paper we use a conservative value from this range of $n_s=13$.

We apply this descriptor in the same way as Arandjelovi\'c and Zisserman did theirs in the BoB method \citep{AranZiss2011}, or indeed a number of authors before them using local texture descriptors \citep{SiviRussEfro+2005}. The set of training descriptors is first clustered, the centre of each cluster defining the corresponding descriptor word. An object is then represented by a histogram of its descriptor words. Since we too do not encode any explicit geometric information between individual descriptors we refer to our representation as a bag of normals (BoN).

\paragraph{Computational advantages.}
An analysis of the computational demands of the BoN and BoB representations readily demonstrates superior efficiency of the proposed BoN. Firstly, the time needed to extract the proposed descriptors from a boundary is dramatically lower than those of Arandjelovi\'c and Zisserman -- approximately 16 times in our implementation\footnote{Matlab, running on an AMD Phenom II X4 965 processor and 8GB RAM.}. The total memory needed to store the extracted descriptors per object is also reduced, to approximately 8\%. Unlike the descriptor of Arandjelovi\'c and Zisserman which is affected by the confounding image information surrounding the boundary, the proposed descriptor describes local boundary shape direction. Thus, the size of the vocabulary of boundary features need not be as large (see Section~\ref{s:eval}). This means that the total storage needed for the representations of all objects in a database is smaller, and their matching faster (the corresponding histograms are shorter).

\vspace{-6pt}\subsection{Object texture representation}\label{ss:text}
The second part of our characterization of an object seeks to describe its texture. Here we adopt the standard representation in the form of a histogram of visual words \citep{NistStew2006,PhilChumIsarSivi+2007}, based on Lowe's SIFT \citep{Lowe2004}. Since we claim no significant novelty here, we only briefly summarize the approach. Firstly, a sparse set of interest points is detected as the well localized \textit{extrema} of a difference of Gaussians scale-space constructed from the original image. A descriptor is extracted at each keypoint from the corresponding $16 \times 16$ pixel patch. The descriptor is computed by concatenating weighted 8-bin histograms of gradients across 16 non-overlapping $4 \times 4$ pixel neighbourhoods of the patch. The descriptors computed across the entire training data set are used to learn a vocabulary of textural words by $K$-means clustering.

\vspace{-6pt}\subsection{Combining textural and shape representations}\label{ss:comb}
As noted earlier, the apparent texture and shape of an object are virtually entirely complementary. This observation motivates a simple fusion rule in the form of a weighted summation of the corresponding histogram differences. If $\Delta h_t$ is the difference between the histograms of textural words of two objects and $\Delta h_s$ the difference between their histograms of shape words, we compute the overall dissimilarity (``distance'') $d$ between two objects as:
\begin{align}
  d = (1-W) \times \Delta h_t + W \times \Delta h_s
\end{align}
The optimal value of the weighting constant $W$ is data dependent. As we illustrated in Section~\ref{s:intro}, in the case of some objects texture is more discriminative than shape and for some it is the other way round. Thus, our aim is to learn the optimal value of $W$ for a specific training data set.

The main challenge of interest to us is the change in the viewpoint from which objects are imaged and which greatly affects their appearance. Since we assume that the system is trained using a single image per object, we estimate the optimal value of the weighting constant using synthetically generated data. Specifically, from each training image we generate a number of geometrically distorted images by applying small random affine warps (in a manner similar to \citep{AranCipo2006c}). The aim of these warps is to approximate the changes in the objects' appearance which would be effected with small out of plane rotations. The optimal value of the weighting constant is then determined by maximizing the difference between inter-class and intra-class distances, the former being computed using the actual training data of different objects and the latter using synthetically generated images.

\vspace{-6pt}\section{Evaluation}\label{s:eval}

\begin{figure*}[htb]
  \centering
  \includegraphics[width=1\textwidth]{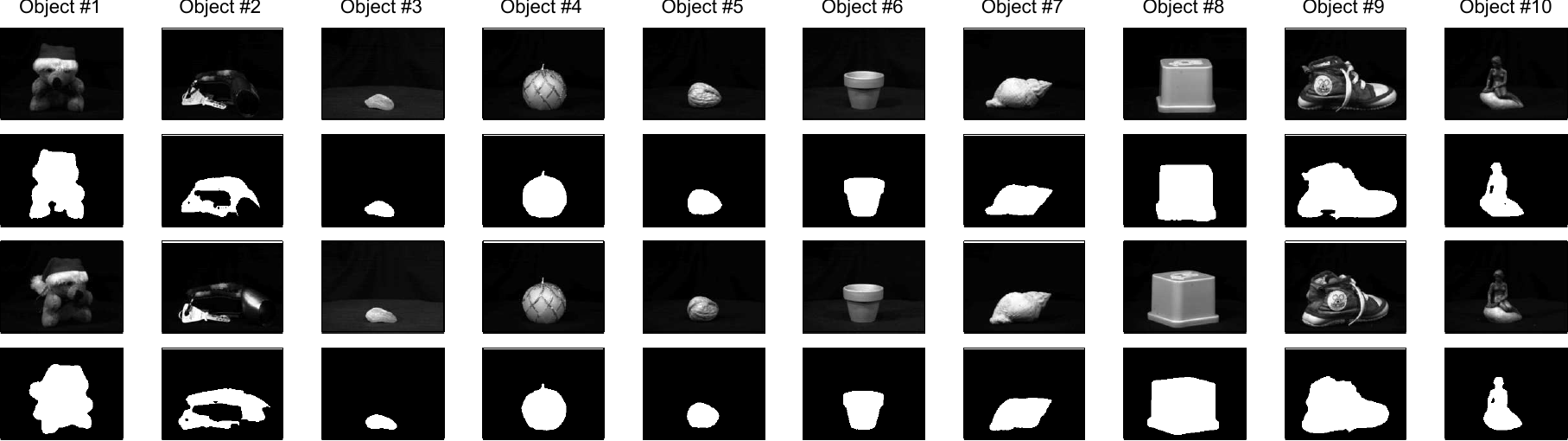}
  \caption{ The first 10 objects in the \emph{Amsterdam Library of Object Images} (ALOI) \citep{GeusBurgSmeu2005} seen from two views 30$^\circ$ apart (first and third row) and the corresponding foreground/background masks, extracted automatically using pixel intensity thresholding.  }
  \label{f:aloi}
\end{figure*}

\paragraph{Data set.}
To evaluate the effectiveness of the proposed algorithm we used the publicly available \emph{Amsterdam Library of Object Images} (ALOI) \citep{GeusBurgSmeu2005}. This data set comprises images of 1000 objects, each imaged from 72 different viewpoints, at successive 5$^{\circ}$ rotations about the vertical axis (i.e.\ yaw changes). We used a subset of this variation, constrained to viewpoint directions of 0--85$^{\circ}$. The objects in the database were imaged in front of a black background, allowing a foreground/background mask to be extracted automatically using simple thresholding. This is illustrated using the first 10 objects in the database in Figure~\ref{f:aloi}. This segmentation was performed by the authors of the database, rather than the authors of this paper. It should be emphasized that the result of aforesaid automatic segmentation is not perfect. Errors were mainly caused by the dark appearance of parts of some objects, as well as shadows. This is readily noticed in Figure~\ref{f:aloi} and in some cases, the deviation from the perfect segmentation result is much greater than that shown and, importantly, of variable extent across different viewpoints.

\paragraph{Parameters.}
We learn the discretized vocabularies of both local textural and shape descriptors using the 1000 images of all objects seen from the nominal 0$^\circ$ viewpoint. We used a 5000 word vocabulary of textural words and a 3000 word vocabulary of shape words. The size of the latter is smaller because the possible variability of shape is inherently less complex than that of textural appearance. Our shape vocabulary is also smaller than that used in the BoB method of Arandjelovi\'c and Zisserman. This is because in our method shape is described directly, rather than implicitly by means of an image based descriptor in the BoB.

\begin{figure*}[htb]
  \centering
  \subfigure[Rank-1]{\includegraphics[width=0.35\textwidth]{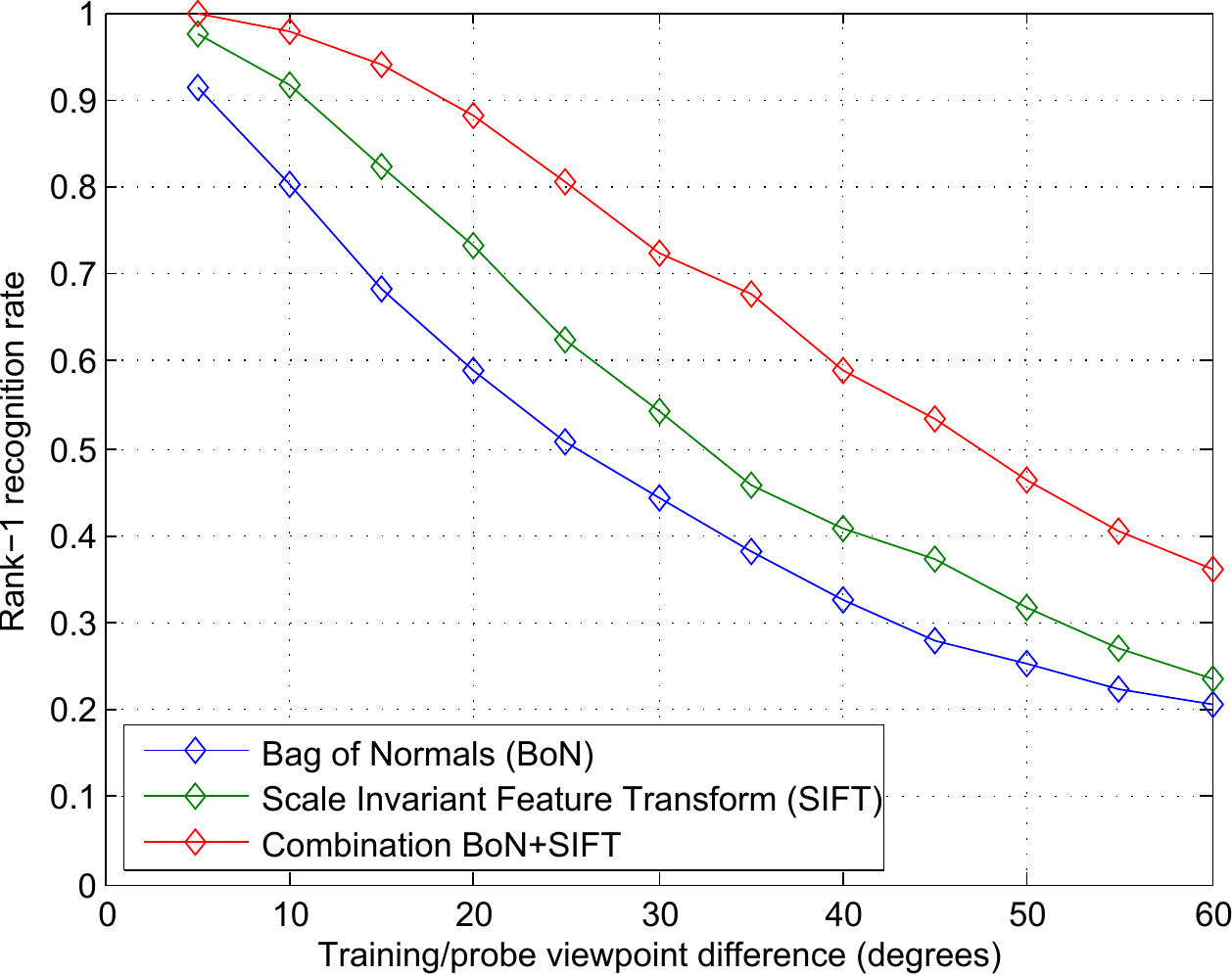}}\hspace{20pt}
  \subfigure[Rank-5]{\includegraphics[width=0.35\textwidth]{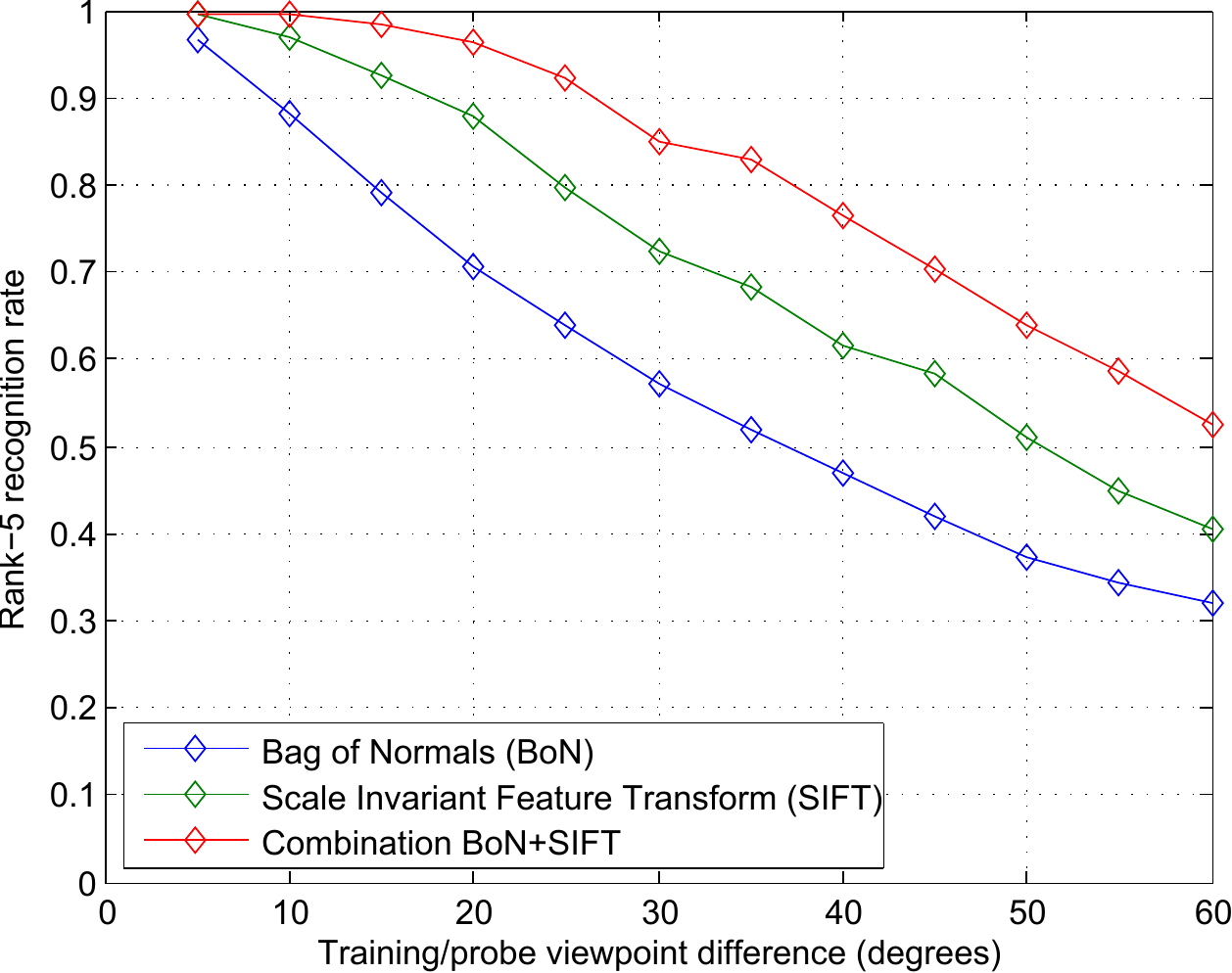}}
  \subfigure[Rank-10]{\includegraphics[width=0.35\textwidth]{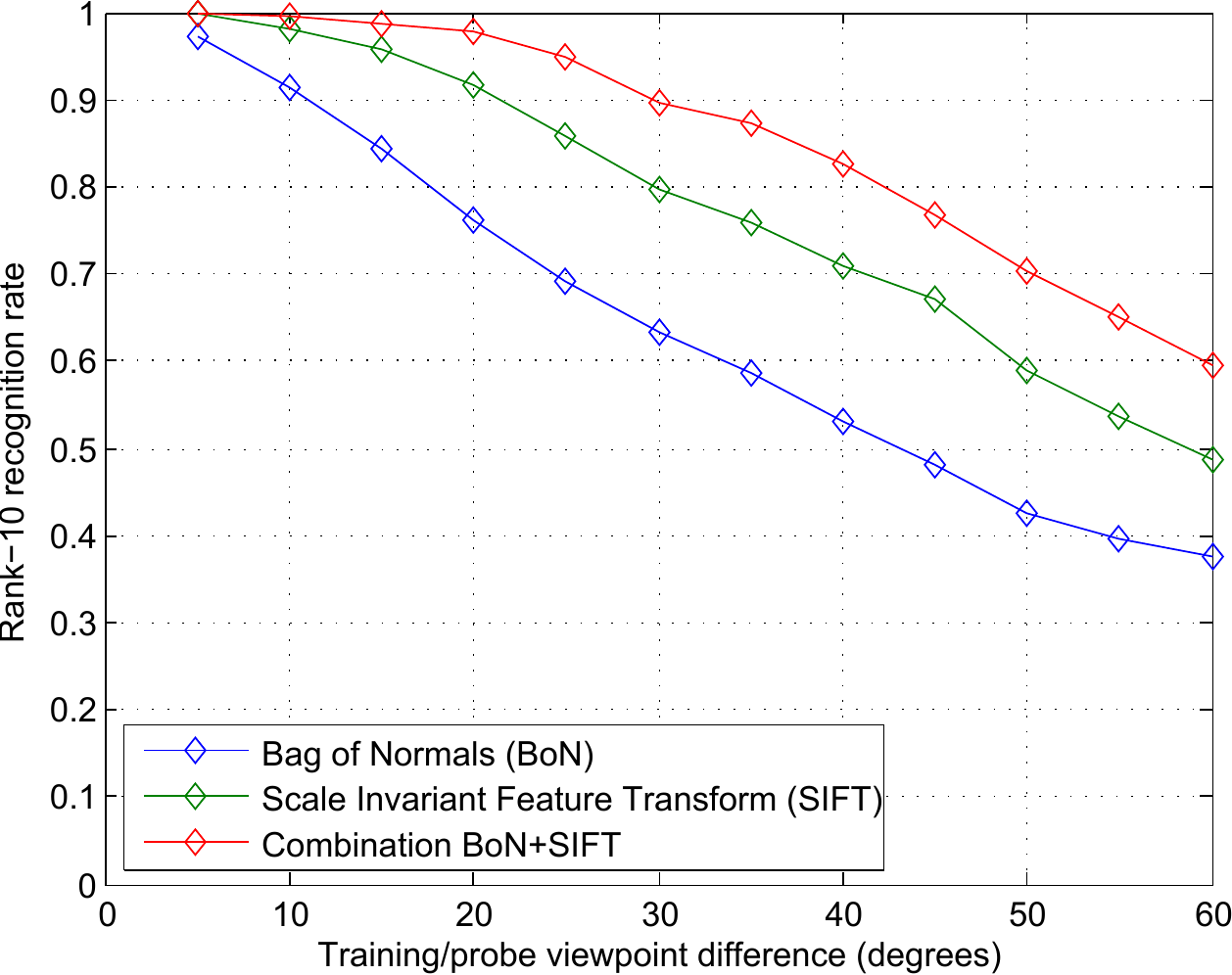}}\hspace{20pt}
  \subfigure[Rank-20]{\includegraphics[width=0.35\textwidth]{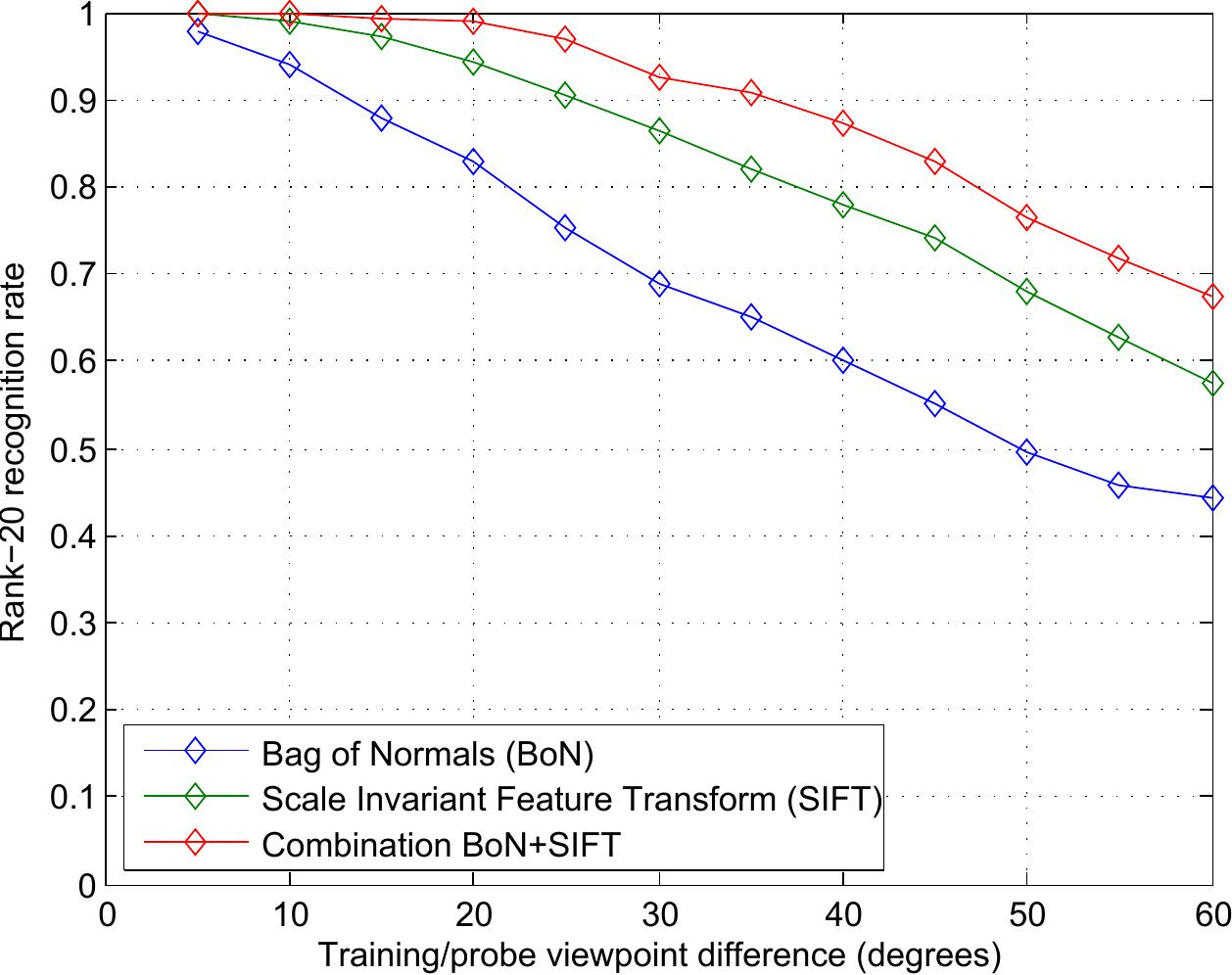}}
  \caption{ The variation of rank-$N$ recognition rate ($N \in \{1,5,10,20\}$) across viewpoint differences of 5--85$^\circ$ between training and probe images. On the ALOI data set, textural SIFT based matching of objects consistently outperforms BoN based shape matching. The proposed combination of textural and shape descriptors significantly outperforms both modalities in isolation. }
  \label{f:res}
\end{figure*}

\paragraph{Results.}
The key results of our evaluation are shown in Figure~\ref{f:res} which plots the variation of the rank-$N$ recognition rate (for $N \in \{1,5,10,20\}$) attained with different object representations as a function of the viewpoint difference between the training image set and query images. Firstly, note that the purely textural representation based on SIFT features outperforms the purely apparent shape based BoN representation on this data set. This is observed consistently across viewpoint changes and different rank thresholds. This suggests that the type of objects in the ALOI data set is closer to the textured extreme of the textured-smooth object continuum described in Section~\ref{s:intro}. Interestingly, the performance difference between purely textural and purely shape based representations is the smallest for rank-1 recognition and increases with the value of $N$. For example, at 30$^\circ$ yaw difference between training and query images, the difference in rank-1, rank-5, rank-10 and rank-20 recognition rates is respectively 10.0\%, 15.4\%, 16.4\% and 17.4\%.

The proposed method of combining textural and shape features also consistently improves matching performance, across viewpoint variations and different rank thresholds. Moreover, the improvement is substantial -- for example, at 20$^\circ$ yaw difference between training and query images the rank-1 error rates of 40.1\% and 26.8\% for purely shape and purely texture based matching is reduced down to 11.9\% only. The improvement is even more significant for rank-5, rank-10 and rank-20 performance, reducing the error rates respectively from 29.5\% and 12.2\% to 3.6\%, from 23.7\% and 8.2\% to 2.2\%, and from 17.2\% and 5.6\% to 1.0\%. It is interesting to note that for a large viewpoint change ($\approx 60 ^\circ$) both purely shape and purely texture based representations perform rather poorly (20.6\% and 23.4\% correct recognition rate respectively); yet despite this the proposed combination of the two modalities nearly doubles the correct recognition rate. This and the previously described results confirm the fundamental premise of the present work that textural and shape modalities contain complementary discriminative information. They also demonstrate the effectiveness of both the proposed shape descriptor that the BoN representation is based upon, and the effectiveness of our method for fusing textural and shape information.

\vspace{-6pt}\section{Summary and conclusions}
In this paper we addressed the problem of matching images of 3D objects across camera viewpoint changes. One of the key premises of our work is that a description of an object should account both for its texture and its apparent shape. The task of matching textured objects has been one of the most active research areas of computer vision in the last decade and has already achieved success in practical applications. In contrast, the handling of smooth objects for which texture based methods fail has only recently attracted some research attention. Our work is the first to investigate the use of both modalities. We adopt a histogram based representation of discretized elementary descriptors: for texture, the elementary descriptor is SIFT based, while for shape we describe a novel local descriptor based on the directions of local boundary normals. In contrast with previous work on apparent shape based matching of 3D objects, we do not describe the entirety of the object's boundary but rather only its salient segments. A method for determining these segments automatically is described and bears some resemblance to interest point detection in 2D images. Lastly, experiments on a large and publicly available database of a diverse set of objects were used to evaluate the proposed methodologies empirically. Our fusion of textural and shape representations was shown to dramatically improve matching performance. The improvement was particularly significant in the case of a large viewpoint change between training and query data, when the correct rank-1 recognition rate was nearly twice that achieved using either of the modalities in isolation.

\small
\bibliographystyle{named}
\bibliography{../../my_bibliography}

\end{document}